\pdfoutput=1

\documentclass[11pt]{article}

\usepackage[final]{acl}

\usepackage{times}
\usepackage{latexsym}
\usepackage{multirow}
\usepackage[table]{xcolor}
\usepackage{booktabs}
\usepackage{graphicx}
\usepackage{array}
\usepackage{setspace}
\usepackage{float}
\usepackage{longtable}
\usepackage{ragged2e}
\usepackage{placeins}
\usepackage{amssymb}

\usepackage[most]{tcolorbox}

\usepackage[T1]{fontenc}

\usepackage[utf8]{inputenc}

\usepackage{microtype}

\usepackage{inconsolata}

\usepackage{graphicx}

\usepackage{amsmath}
\usepackage{amsthm}
\usepackage{booktabs}
\usepackage{algorithm}
\usepackage{algorithmic}

\title{VecCISC: Improving Confidence-Informed Self-Consistency with Reasoning Trace Clustering and Candidate Answer Selection}

\author{
James Petullo\textsuperscript{*} \and Sonny George \and Dylan Cashman \and Nianwen Xue \\
Computer Science Department, Brandeis University \\
\texttt{jamespetullo@brandeis.edu} \\
\textsuperscript{*}Corresponding author
}

\begin{document}
\maketitle
\begin{abstract}
A standard technique for scaling inference-time reasoning is Self-Consistency, whereby multiple candidate answers are sampled from an LLM and the most common answer is selected. More recently, it has been shown that weighted majority voting (e.g. Confidence-Informed Self Consistency (CISC)), which assigns a confidence value to each candidate answer and chooses the answer with the largest accumulated score, tends to be more accurate on a wide range of popular benchmarks. In practice, weighted majority voting necessitates calling a critic LLM on each candidate’s reasoning trace to produce the answer’s confidence score. This secondary series of LLM calls greatly increases the overhead and cost of weighted majority voting, despite its potential performance benefits. To reduce this expense, we propose VecCISC, a lightweight, adaptive framework that uses a measure of semantic similarity to filter reasoning traces that are semantically equivalent to others, degenerate, or hallucinated, thus decreasing the number of candidate answers that must be evaluated by the critic. To ensure adequate experimental thoroughness, we evaluate VecCISC on five challenging, widely-adopted datasets spanning the domains of mathematics, chemistry, biology, commonsense reasoning, and the humanities. Our results demonstrate that VecCISC reduces the total token usage by 47\%, while maintaining or exceeding the accuracy of CISC.
\end{abstract}

\section{Introduction}
Scaling inference-time reasoning has become an effective and popular approach to improving the performance of LLMs on reasoning tasks \cite{2, 3, 4, 5}. At its core, scaling inference-time reasoning involves sampling multiple outputs from an LLM and, from them, producing a final answer. The underlying assumption is that a larger set of diverse reasoning traces will lead to a greater number of potential answers, thus improving coverage of the solution search space. The foundational technique for inference scaling is Self-Consistency \cite{6}, whereby an LLM is called multiple times for a given prompt, and the most common answer is returned. Since these methods can be costly, efforts have been made to decrease sampling costs while maintaining accuracy \cite{8}. These approaches include early stopping \cite{9, 10, 11}, mixture of models \cite{1, 14}, and query routing \cite{13}. Recently, it has been shown that confidence-informed Self-Consistency (CISC) \cite{12} can be utilized in conjunction with weighted majority voting to improve the accuracy of Self-Consistency. CISC, and the broader concept of LLM confidence calibration, belongs to the \textbf{“think twice” paradigm} \cite{15}, whereby an LLM first generates an answer and an associated reasoning trace, and then a critic LLM is prompted to assess the answer-trace pair. The critic produces a score that reflects its confidence that the answer is correct. These scores are then used as part of a weighted majority vote to select the final answer. Throughout this paper, we use the term “think twice” to refer to any method that necessitates calling a critic for confidence estimation.

However, a significant drawback to “think twice” approaches is that the inference overhead increases significantly, as each sample requires at least one additional LLM call, often with a token-heavy prompt. Furthermore, naive CISC makes no distinction between hallucinated, degenerate samples and higher quality outputs, running the critic LLM on all regardless. 

As such, we introduce VecCISC, a flexible, conceptually straightforward candidate answer aggregation framework that reduces the number of samples needed to be evaluated by the critic LLM through clustering candidate answers via embeddings of their corresponding reasoning traces. By adding only minimal overhead through the application of an embedding model, our clustering approach filters redundant traces that possess the same semantic meaning (see Appendix \ref{sec:cs1}) and can aid in isolating answers that contain flawed, hallucinated, and degenerate reasoning (see Appendix \ref{sec:cs2}). After clustering, a representative answer from each cluster is selected to be passed to the critic LLM, thus significantly reducing the number of calls that have to be made to the latter. In sum, our contributions are as follows:

\begin{itemize}
    \item We introduce VecCISC, a novel method to reduce the cost of the “think twice” paradigm by utilizing reasoning trace clustering to eliminate redundant and low quality samples from the pool to be analyzed;
    \item We empirically demonstrate the effectiveness of our approach by performing extensive experiments on a range of models and datasets, with VecCISC achieving an average total token budget reduction of 47\% across the entire inference pipeline, all while maintaining and even exceeding the performance of CISC;
    \item We compare the performance of VecCISC with two different clustering approaches against random candidate selection of $K$ samples, demonstrating the efficacy of our proposed clustering approach;
    \item We will release our code publicly to support further research.
\end{itemize}

\section{Methodology}
\begin{figure*}
\centering
\includegraphics[width=1\textwidth]{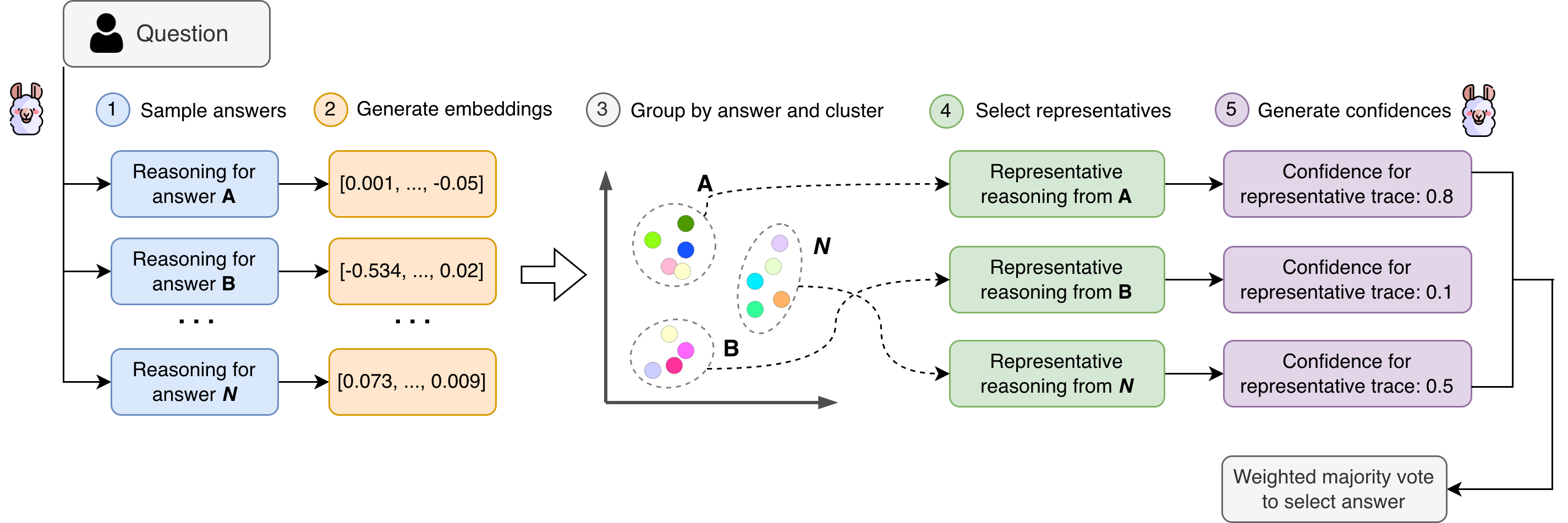}
\caption{Overview of the VecCISC pipeline. Embeddings of the sampled reasoning traces are clustered within each unique answer returned from the LLM. A representative trace is chosen from each cluster, and these representatives are then used to generate the confidence scores for the answers. Lastly, weighted majority vote is performed to choose the final answer.
}
\label{fig:pipeline_overview}
\end{figure*}

\begin{figure*}
\centering
\includegraphics[width=0.7\textwidth]{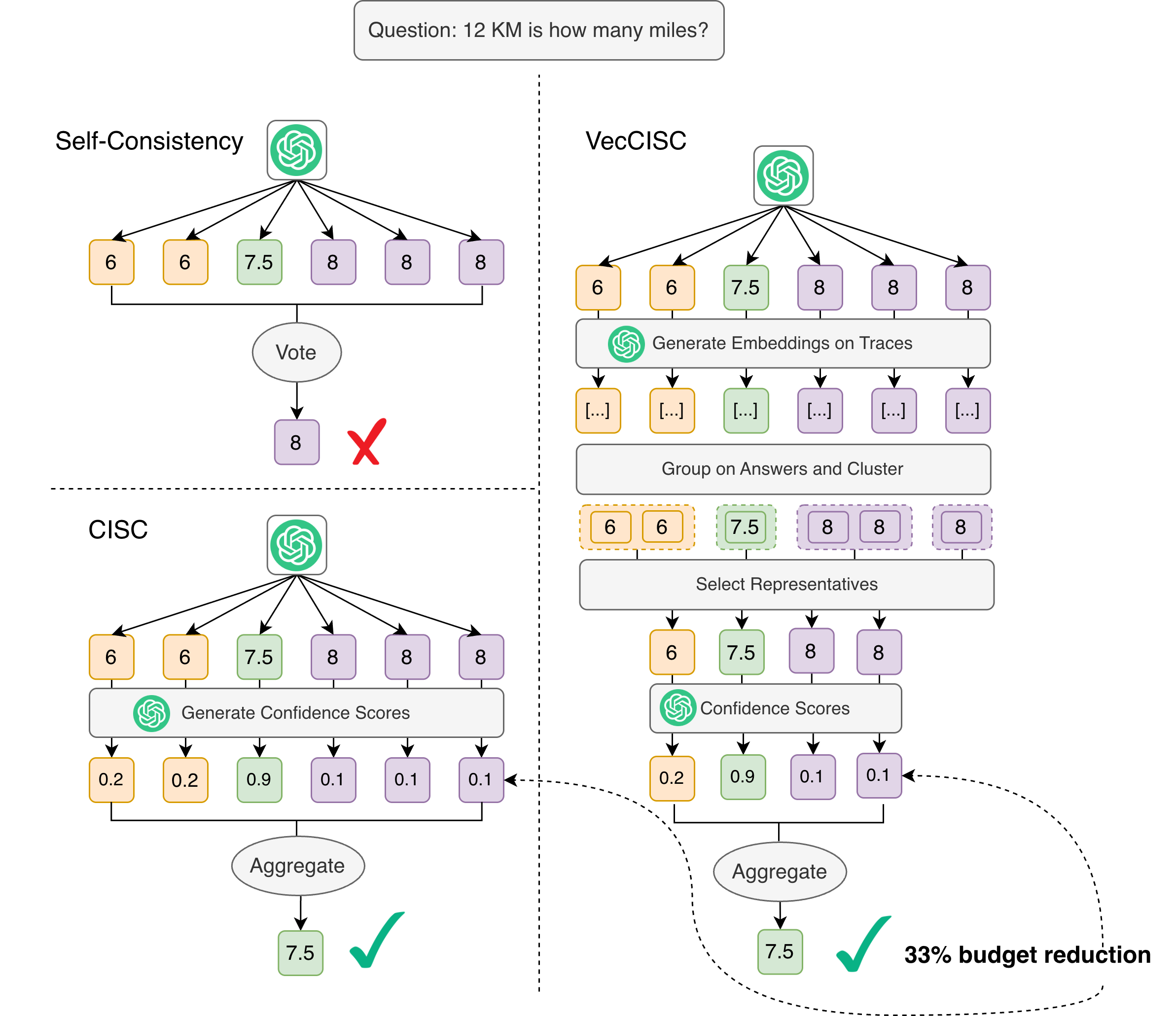}
\caption{Comparison of VecCISC to Self-Consistency (SC) and CISC. While CISC represents an improvement upon SC via incorporation of confidence scores and weighted majority voting, the inference overhead costs of CISC increase twofold over SC. By incorporating language embeddings, clustering, and representative trace selection, the confidence scorer $LLM_{critic}$ can be run on a smaller set of higher quality reasoning traces while still preserving relative performance.
}
\label{fig:comparison}
\end{figure*}

This section details each step in the VecCSIC pipeline, including embedding generation from answer reasoning traces, clustering, and candidate answer selection. The overall procedure is summarized in Figure \ref{fig:pipeline_overview}, and a comparison of VecCISC’s methodology to that of Self-Consistency and CISC can be found in Figure \ref{fig:comparison}.

\subsection{Sampling}

Given a sampling budget $n$, a large language model $LLM_{gen}$, and a question $q$, we first sample $n$ reasoning trace-answer pairs from $LLM_{gen}$:

\begin{equation}
{(r_1, a_1), … (r_n, a_n)} \sim LLM_{gen}(q)
\end{equation}

The samples are then added to the answer candidate pool $P$:

\begin{equation}
P = \{(r_1, a_1), … (r_n, a_n)\}
\end{equation}

\subsection{Embedding Generation}

For each reasoning trace in the pool, a text embedding is produced via an embedding model $Emb$:

\begin{equation}
\begin{aligned}
P = \{ (r_i, \mathrm{Emb}(r_i), a_i) \mid (r_i, a_i) \in P \},\\ \mathrm{Emb}(r_i) \in \mathbb{R}^d
\end{aligned}
\end{equation}

Where $d$ is the embedding dimension.\\

These embeddings enable the pipeline to capture semantic details in each reasoning trace that allow for clustering and outlier detection in subsequent steps.

\subsection{Grouping By Answer}

To preserve the set of distinct candidate answers, we group embeddings by answer before clustering. That is, for each unique answer $a$ present in the sampled outputs:

\begin{equation}
\begin{aligned}
G_a = \{ (r_i, \mathrm{Emb}(r_i)) \mid & \, (r_i, \mathrm{Emb}(r_i), a_i) \in P, \\
& \, a_i = a \}
\end{aligned}
\end{equation}

Where $G_a$ is the group of embedded reasoning traces that lead to answer $a$.

\subsection{Clustering}

Next, we apply a clustering algorithm $Cluster$ to each reasoning trace group $G_a$:

\begin{equation}
\begin{aligned}
\text{CL}_a = \text{Cluster}\Big( 
& \{ \mathrm{Emb}(r_i) \mid (r_i, \mathrm{Emb}(r_i)) \in G_a \}, \\
& \min(K, |G_a|) 
\Big)
\end{aligned}
\end{equation}

Where $K$ is a hyperparameter controlling the number of clusters to be formed. If $K$ exceeds the number of samples in $G_a$, then $|G_a|$ clusters are created instead. In our experiments, we analyzed KMeans and Hierarchical Agglomerative (HAC) clustering, as both methods allow for a distinct number of clusters to be specified. For both clustering algorithms, we performed a grid search on a holdout set from each $(dataset, model)$ combination to choose $K$. The hyperparameters selected and further technical details are listed in Appendix A.

We choose KMeans and HAC over Density-Based Spatial Clustering of Applications with Noise (DBSCAN) \cite{32}, another popular choice for clustering embeddings, as DBSCAN requires a distance threshold $D$ to define clusters. $D$ specifies the maximum distance between two vectors $V_1$ and $V_2$ for them to be considered part of the same neighborhood. Thus, $V_1$ and $V_2$ are clustered together if $dist(V_1, V_2) \le D$. In high-dimensional language model embedding spaces, small changes to $D$ can drastically alter cluster assignments, making threshold selection both difficult and prone to producing suboptimal clusters.

\subsection{Cluster Representative Selection}
Upon creation of the clusters, we then pick a representative from each cluster $C_i \in CL_a$. To do so, we first compute the centroid $u_i$ of $C_i$:

\begin{equation}
U_i = \frac{1}{|C_i|} \sum_{e \in C_i} e
\end{equation}

Finally, we choose the reasoning trace $r_{rep, i} \in C_i$ that is closest to the centroid:

\begin{equation}
r_{\text{rep}, i} = \arg\min_{r \in C_i} \, \text{cos\_sim}(\mathrm{Emb}(r), u_i)
\end{equation}

Here, we use the cosine similarity function as our definition of distance, as it emphasizes the angle $\theta$ between vectors instead of the magnitude, thus making it more suitable for high-dimensional spaces requiring the preservation of semantic overlap \cite{10}. Cosine similarity is defined in Equation \ref{eq:cosine}:

\begin{equation}
\text{cos\_sim}(\mathrm{emb_1}, \mathrm{emb_2}) 
= \frac{\mathrm{emb_1} \cdot \mathrm{emb_2}}
       {\|\mathrm{emb_1}\| \cdot \|\mathrm{emb_2}\|}
\label{eq:cosine}
\end{equation}

Selecting the reasoning trace closest to the centroid follows from the observation that the least semantically deviant trace would be the least likely to contain anomalous reasoning errors. At the end of the selection process, each answer now possesses a set of representative traces:

\begin{equation}
\begin{aligned}
\text{Rep}_a &= \{ r_{\text{rep},1}, \dots, r_{\text{rep},k} \}, \\
\text{where } k &= |\text{Clusters}_a|, \quad k \leq |G_a|
\end{aligned}
\end{equation}

\subsection{Confidence Scoring}

For each unique answer $a$ and its associated representative traces $r_i \in Rep_a$, we derive a new prompt query $q’_i = [instruction] \circ q \circ r_i \circ a$, where $\circ$ is the concatenation operator and $[instruction]$ prompts the critic LLM to reflect on the original question, reasoning trace, and answer, and produce a confidence score in the range of 0 to 1:

\begin{equation}
c_{a,i} = LLM_{critic}(q’_i)
\end{equation}

Following \cite{12}, we normalize each confidence score using softmax:

\begin{equation}
\hat{c}_{a,i} = \frac{\exp\left( \frac{\text{c}_{a,i}}{T} \right)}
               {\sum_{j=1}^{K} \exp\left( \frac{\text{c}_{a,j}}{T} \right)}
\end{equation}

Where $T$ is a tunable hyperparameter of softmax that controls the smoothness of the normalization distribution. As with the cluster number $K$, we perform a grid search to find $T$. Please see Appendix A for more details and $T$ values from our experiments.

We select the final answer by using a weighted majority vote:

\begin{equation}
A_{\text{final}} = \arg\max_{a} \sum_{(a_j, \hat{c}_{a,j}) \in P} 
1[a_j = a] \cdot \hat{c}_{a,j}
\end{equation}

\section{Experiments}
\subsection{Datasets}

We evaluated the performance of VecCISC on five popular QA datasets:

\begin{itemize}
    \item \textbf{AQuA\_Rat} \cite{17}: A multiple choice QA dataset consisting of algebraic word problems. We sampled 1000 questions from the training set.
    
    \item \textbf{CommonsenseQA} \cite{18}: A dataset that requires commonsense reasoning capabilities to solve problems across a spectrum of different domains. We sampled 1000 problems from the validation set.

    \item \textbf{ARC-Challenging} \cite{19}: A subset of the most challenging questions in the original ARC dataset, including grade-school level multiple choice science questions. We sampled 1000 problems from the test set.

    \item \textbf{MMLU-Pro} \cite{20}:  A more challenging set of questions from the original MMLU dataset, designed to assess reasoning faculty across disciplines including math, science, economics, and history. We sampled 1000 problems from the test set.

    \item \textbf{GPQA} \cite{21}: A multiple-choice dataset of very challenging questions crafted by experts in biology, physics, and chemistry. We used all 546 questions in the dataset.
    
\end{itemize}

\subsection{Models}
For our experiments, we selected five widely used open and closed source LLMs: GPT-4o mini \cite{22}, Llama 3.1 8B \cite{23}, Llama 3.3 70B Versatile \cite{31}, Qwen 2.5 7B \cite{24}, and Mistral 7B v0.1 \cite{25}. Model hyperparameters can be found in Appendix A.

\subsection{Embeddings}
For all experiments, we used OpenAI’s text-embedding-3-small model \cite{26} for the embedding model $Emb$. This model was chosen to balance cost and general purpose applicability across a plethora of different reasoning tasks. As our goal is to design a lightweight, adaptive framework that will be performant regardless of reasoning trace composition, text-embedding-3-small was most suitable for both our task’s scope and budget limitations.

\subsection{Baselines}
We compare VecCISC to both standard Self-Consistency and CISC. In particular, we apply three distinct flavors of methodology: \textbf{VecCISC (random)}, \textbf{VecCISC + KMeans} and \textbf{VecCISC + HAC}. Instead of clustering, \textbf{VecCISC (random)} randomly samples $K$ reasoning traces from each answer-grouped set, $G_a$, thus acting as an ablation for our method. All prompts were zero-shot. For the sake of experimental validity and to ensure a proper apples-to-apples comparison, we reimplemented both Self-Consistency and CISC, executing them on the same set of sampled questions used to evaluate the various VecCISC configurations. Following \cite{12}, we selected verbal confidence as the confidence extraction method, whereby the critic LLM is asked to rate its confidence in the range of 0 to 1.

\subsection{Metrics}

\subsubsection{Budget Call Reduction}
We report the reduction in the number of calls made to $LLM_{critic}$ with VecCISC, as opposed to CISC:

\begin{equation}
\text{Reduction} = 100 \times \frac{\# \text{ of critic calls in VecCISC}}{\# \text{ of calls in CISC}}
\end{equation}

We computed the reduction for each $(dataset, model)$ combination by running our pipeline on each question in the dataset with its corresponding model and recorded the average across all questions and experiments.

\subsubsection{Token Usage Reduction}

In addition, we report the reduction in token usage exhibited by VecCISC as opposed to CISC. Within VecCISC, we compare two cluster representative selection strategies: our proposed minimum centroid (min-centroid) approach and random cluster value selection (rand-trace), the latter acting as an ablation of the former.

\subsubsection{Accuracy}
Lastly, we report the accuracy attained by VecCISC and its variants, CISC, and Self-Consistency (SC) on each dataset and model. KMeans clustering and random trace sampling are non-deterministic, thus, we ran each method 10 times for every $(dataset, model)$ combination and reported both the best and average.

\section{Results and Discussion}

Our budget reduction results can be found in Tables \ref{tab:llm_budget_reductions_kmeans} and \ref{tab:llm_budget_reductions_hac}, while the token reduction results can be found in Tables \ref{tab:kmeans_token_reduction} and \ref{tab:hac_token_reduction}. In addition, performance results can be found in Table \ref{tab:performance}.

\begin{table*}[p]
\centering
\renewcommand{\arraystretch}{1.2}
\setlength{\tabcolsep}{4pt}
\begin{tabular}{llccccc}
\hline
\textbf{Dataset} & \textbf{Budget} & \textbf{GPT-4o mini} & \textbf{Llama 3.1 8B} & \textbf{Qwen2.5 7B} & \textbf{Mistral 7B}  & \textbf{Llama3.3 70B} \\
\hline
    
\multirow{1}{*}{AQuA-RAT}
& 20 & 5.660{\scriptsize  (-71.72\%) } & 13.45{\scriptsize  (-32.77\%) } & 14.78{\scriptsize  (-26.09\%) } & 17.05{\scriptsize  (-14.74\%) } & 2.600{\scriptsize  (-86.99\%) } \\
\hline
\multirow{1}{*}{CommonsenseQA}
& 20 & 13.81{\scriptsize  (-30.95\%) } & 12.70{\scriptsize  (-36.50\%) } & 12.45{\scriptsize  (-37.76\%) } & 15.54{\scriptsize  (-22.31\%) } & 11.79{\scriptsize  (-41.02\%) } \\
\hline
\multirow{1}{*}{ARC-Challenging}
& 20 & 13.34{\scriptsize  (-33.29\%) } & 14.80{\scriptsize  (-26.00\%) } & 10.94{\scriptsize  (-45.30\%) } & 15.65{\scriptsize  (-21.73\%) } & 13.31{\scriptsize  (-33.43\%) } \\
\hline
\multirow{1}{*}{MMLU-Pro}
& 20 & 12.28{\scriptsize  (-38.61\%) } & 14.81{\scriptsize  (-25.94\%) } & 12.81{\scriptsize  (-35.93\%) } & 17.47{\scriptsize  (-12.66\%) } & 13.18{\scriptsize  (-34.10\%) } \\
\hline
\multirow{1}{*}{GPQA}
& 20 & 10.56{\scriptsize  (-47.19\%) } & 13.88{\scriptsize  (-30.57\%) } & 16.41{\scriptsize  (-17.93\%) } & 10.03{\scriptsize  (-49.86\%) } & 17.27{\scriptsize  (-13.63\%) } \\
\hline

\end{tabular}
\caption{\textbf{VecCISC + KMeans} confidence evaluator call budget reductions, averaged across 10 runs. \textbf{min-centroid} cluster candidate selection was used for all experiments. \textbf{VecCISC + KMeans} records a 34.68\% reduction in $LLM_{critic}$ calls.}
\label{tab:llm_budget_reductions_kmeans}
\end{table*}

\begin{table*}[p]
\centering
\renewcommand{\arraystretch}{1.2}
\setlength{\tabcolsep}{4pt}
\begin{tabular}{llccccc}
\hline
\textbf{Dataset} & \textbf{Budget} & \textbf{GPT-4o mini} & \textbf{Llama 3.1 8B} & \textbf{Qwen2.5 7B} & \textbf{Mistral 7B}  & \textbf{Llama3.3 70B} \\
\hline
    
\multirow{1}{*}{AQuA-RAT}
& 20 & 13.67{\scriptsize  (-31.64\%) } & 13.45{\scriptsize  (-32.77\%) } & 14.78{\scriptsize  (-26.09\%) } & 17.05{\scriptsize  (-14.74\%) } & 8.100{\scriptsize  (-59.52\%) } \\
\hline
\multirow{1}{*}{CommonsenseQA}
& 20 & 11.89{\scriptsize  (-40.56\%) } & 12.70{\scriptsize  (-36.50\%) } & 15.23{\scriptsize  (-23.83\%) } & 15.54{\scriptsize  (-22.31\%) } & 11.79{\scriptsize  (-41.02\%) } \\
\hline
\multirow{1}{*}{ARC-Challenging}
& 20 & 10.39{\scriptsize  (-48.07\%) } & 14.80{\scriptsize  (-26.00\%) } & 10.94{\scriptsize  (-45.30\%) } & 15.65{\scriptsize  (-21.73\%) } & 13.31{\scriptsize  (-33.43\%) } \\
\hline
\multirow{1}{*}{MMLU-Pro}
& 20 & 12.28{\scriptsize  (-38.61\%) } & 15.56{\scriptsize  (-22.18\%) } & 12.81{\scriptsize  (-35.93\%) } & 18.02{\scriptsize  (-9.890\%) } & 11.29{\scriptsize  (-43.55\%) } \\
\hline
\multirow{1}{*}{GPQA}
& 20 & 10.56{\scriptsize  (-47.19\%) } & 16.00{\scriptsize  (-20.02\%) } & 16.41{\scriptsize  (-17.93\%) } & 18.81{\scriptsize  (-5.940\%) } & 17.95{\scriptsize  (-10.24\%) } \\
\hline

\end{tabular}
\caption{\textbf{VecCISC + HAC} confidence evaluator call budget reductions, averaged across 10 runs. \textbf{min-centroid} cluster candidate selection was used for all experiments. \textbf{VecCISC + HAC} records a 30.2\% reduction in $LLM_{critic}$ calls.}
\label{tab:llm_budget_reductions_hac}
\end{table*}

\begin{table*}[p]
\centering
\resizebox{\textwidth}{!}{
\begin{tabular}{l|l|cccccc}
\toprule
\textbf{Dataset} & \textbf{Method} & \textbf{GPT 4o-mini} & \textbf{Llama3.1 8B} & \textbf{Qwen2.5 7B} & \textbf{Mistral 7B}  & \textbf{Llama3.3 70B} \\ \midrule

\multirow{3}{*}{AQuA-RAT}
& CISC & 13042.0 & 15480.9 & 11674.3 & 9317.2 & 14448.1 \\
& \cellcolor{gray!20} VecCISC (rand-trace) & \cellcolor{gray!20}\textbf{8914.5} & \cellcolor{gray!20}\underline{10500.6} & \cellcolor{gray!20}\underline{8566.6} & \cellcolor{gray!20}\underline{7652.0} & \cellcolor{gray!20}\underline{6289.3} \\
& \cellcolor{gray!20} VecCISC (min-centroid) & \cellcolor{gray!20}\underline{8915.4} & \cellcolor{gray!20}\textbf{10483.7} & \cellcolor{gray!20}\textbf{8562.6} & \cellcolor{gray!20}\textbf{7649.6} & \cellcolor{gray!20}\textbf{6280.7} \\
\midrule
\multirow{3}{*}{CommonsenseQA}
& CISC & 11086.2 & 10932.4 & 10629.9 & 7691.9 & 13108.5 \\
& \cellcolor{gray!20} VecCISC (rand-trace) & \cellcolor{gray!20}\underline{6546.9} & \cellcolor{gray!20}\textbf{6611.4} & \cellcolor{gray!20}\textbf{8044.4} & \cellcolor{gray!20}\underline{5663.9} & \cellcolor{gray!20}\textbf{7747.4} \\
& \cellcolor{gray!20} VecCISC (min-centroid) & \cellcolor{gray!20}\textbf{6546.7} & \cellcolor{gray!20}\underline{6612.5} & \cellcolor{gray!20}\underline{8044.7} & \cellcolor{gray!20}\textbf{5663.7} & \cellcolor{gray!20}\underline{7748.2} \\
\midrule
\multirow{3}{*}{ARC-Challenging}
& CISC & 12245.4 & 12292.1 & 11602.7 & 8764.0 & 14715.5 \\
& \cellcolor{gray!20} VecCISC (rand-trace) & \cellcolor{gray!20}\underline{6367.2} & \cellcolor{gray!20}\underline{8913.2} & \cellcolor{gray!20}\textbf{6327.2} & \cellcolor{gray!20}\textbf{6733.4} & \cellcolor{gray!20}\underline{9863.1} \\
& \cellcolor{gray!20} VecCISC (min-centroid) & \cellcolor{gray!20}\textbf{6365.6} & \cellcolor{gray!20}\textbf{8912.3} & \cellcolor{gray!20}\underline{6327.5} & \cellcolor{gray!20}\underline{6734.1} & \cellcolor{gray!20}\textbf{9856.5} \\
\midrule
\multirow{3}{*}{MMLU-Pro}
& CISC & 16975.1 & 18496.6 & 16474.9 & 13926.1 & 20810.4 \\
& \cellcolor{gray!20} VecCISC (rand-trace) & \cellcolor{gray!20}\underline{10173.7} & \cellcolor{gray!20}\underline{13616.7} & \cellcolor{gray!20}\underline{10060.1} & \cellcolor{gray!20}\textbf{11140.1} & \cellcolor{gray!20}\underline{11840.2} \\
& \cellcolor{gray!20} VecCISC (min-centroid) & \cellcolor{gray!20}\textbf{10172.4} & \cellcolor{gray!20}\textbf{13614.9} & \cellcolor{gray!20}\textbf{10059.9} & \cellcolor{gray!20}\underline{11143.5} & \cellcolor{gray!20}\textbf{11834.7} \\
\midrule
\multirow{3}{*}{GPQA}
& CISC & 18594.2 & 19659.5 & 17937.0 & 14324.6 & 25235.3 \\
& \cellcolor{gray!20} VecCISC (rand-trace) & \cellcolor{gray!20}\textbf{9574.2} & \cellcolor{gray!20}\underline{15224.9} & \cellcolor{gray!20}\textbf{14419.0} & \cellcolor{gray!20}\underline{13262.6} & \cellcolor{gray!20}\textbf{22386.5} \\
& \cellcolor{gray!20} VecCISC (min-centroid) & \cellcolor{gray!20}\underline{9577.6} & \cellcolor{gray!20}\textbf{15211.4} & \cellcolor{gray!20}\underline{14422.6} & \cellcolor{gray!20}\textbf{13260.5} & \cellcolor{gray!20}\underline{22389.4} \\

\bottomrule
\end{tabular}
}
\caption{\textbf{VecCISC + KMeans} token reductions with cluster candidate selection variants, averaged across 10 runs. In each column, the best results are in \textbf{bold}, and the second-best results are \underline{underlined}.\textbf{VecCISC + KMeans} records a 36.2\% reduction in tokens expended on $LLM_{critic}$ calls.}
\label{tab:kmeans_token_reduction}
\end{table*}

\subsection{Call Budget Reduction}

 \textbf{VecCISC + KMeans} and \textbf{VecCISC + HAC}  significantly reduces the number of calls to $LLM_{critic}$, while achieving comparable or improved performance over CISC. Across all models and datasets, \textbf{VecCISC + KMeans} averages a 34.68\% reduction in the number of critic calls (Table \ref{tab:llm_budget_reductions_kmeans}), while \textbf{VecCISC + HAC} averages a 30.2\% reduction in the number of critic calls (Table \ref{tab:llm_budget_reductions_hac}). Across the entire pipeline, including both the Self-Consistency sampling and CISC component, \textbf{VecCISC + KMeans} averages a 17.34\% total LLM call reduction and \textbf{VecCISC + HAC} averages a 15.1\% total call reduction.

 \subsection{Token Usage Reduction}

By reducing the number of critic calls made, VecCISC variants reduce the number of tokens utilized by $LLM_{critic}$. Our proposed cluster representative selection (min-centroid) records the lowest token usage rates as opposed to random representative selection (rand-trace) on 15 of the 25 model-dataset combinations (60\%) on \textbf{VecCISC + KMeans}, as shown in Table \ref{tab:kmeans_token_reduction}. On \textbf{VecCISC + HAC}, min-centroid records the lowest token usage rates on 17 of the 25 model-dataset combinations (68\%), as shown in Table \ref{tab:hac_token_reduction}. These results demonstrate that our min-centroid approach leads to the selection of higher quality, lower-token reasoning traces as opposed to purely random reasoning trace selection from each cluster. On average, \textbf{VecCISC + KMeans} reduces critic token usage by 36.2\% ( Table \ref{tab:kmeans_token_reduction}), while \textbf{VecCISC + HAC} records a 31.69\% token usage reduction (Table \ref{tab:hac_token_reduction}). $LLM_{critic}$ calls are the most token-heavy of the entire pipeline, accounting for 77\% of all token usage, thus, at scale, these reductions represent significant cost savings. Averaging token consumption across the entire pipeline, both variants of our proposed methodology reduce token usage by 47\% \footnote{To ensure fair approximations across all models tested, we computed the number of tokens used in each prompt as $length(prompt)/ 4$.}.

\subsection{Accuracy}

 As shown in in Table \ref{tab:performance}, \textbf{VecCISC + KMeans} and \textbf{VecCISC + HAC} consistently outperform CISC and Self-Consistency across nearly all models and datasets. \textbf{VecCISC + HAC} records the best average result on all $(dataset, model)$ combinations except for (AQuA\_RAT, Mistral 7B) and (CommonsenseQA, GPT 4o-mini), where it scores the second highest. On (AQuA\_RAT, Mistral 7B) and (CommonsenseQA, GPT 4o-mini), \textbf{VecCISC + KMeans} performs the best. As expected, \textbf{VecCISC (random)}, which randomly samples candidate reasoning traces from each answer group $G_a$ as opposed to clustering, significantly underperforms both CISC and Self-Consistency, thus demonstrating the efficacy of clustering reasoning traces to find an ideal representative candidate.

\begin{table*}[p]
\centering
\resizebox{\textwidth}{!}{
\begin{tabular}{l|l|cccccc}
\toprule
\textbf{Dataset} & \textbf{Method} & \textbf{GPT 4o-mini} & \textbf{Llama3.1 8B} & \textbf{Qwen2.5 7B} & \textbf{Mistral 7B}  & \textbf{Llama3.3 70B} \\ \midrule

\multirow{3}{*}{AQuA-RAT}
& CISC & 13042.0 & 15480.9 & 11674.3 & 9317.2 & 14448.1 \\
& \cellcolor{gray!20} VecCISC (rand-trace) & \cellcolor{gray!20}\textbf{8914.0} & \cellcolor{gray!20}\underline{10499.9} & \cellcolor{gray!20}\underline{8565.2} & \cellcolor{gray!20}\underline{7652.3} & \cellcolor{gray!20}\underline{6291.0} \\
& \cellcolor{gray!20} VecCISC (min-centroid) & \cellcolor{gray!20}\underline{8915.4} & \cellcolor{gray!20}\textbf{10483.7} & \cellcolor{gray!20}\textbf{8562.6} & \cellcolor{gray!20}\textbf{7649.6} & \cellcolor{gray!20}\textbf{6280.7} \\
\midrule
\multirow{3}{*}{CommonsenseQA}
& CISC & 11086.2 & 10932.4 & 10629.9 & 7691.9 & 13108.5 \\
& \cellcolor{gray!20} VecCISC (rand-trace) & \cellcolor{gray!20}\textbf{6546.6} & \cellcolor{gray!20}\textbf{6611.0} & \cellcolor{gray!20}\underline{8044.9} & \cellcolor{gray!20}\underline{5664.7} & \cellcolor{gray!20}\textbf{7746.3} \\
& \cellcolor{gray!20} VecCISC (min-centroid) & \cellcolor{gray!20}\underline{6546.7} & \cellcolor{gray!20}\underline{6612.5} & \cellcolor{gray!20}\textbf{8044.7} & \cellcolor{gray!20}\textbf{5663.7} & \cellcolor{gray!20}\underline{7748.2} \\
\midrule
\multirow{3}{*}{ARC-Challenging}
& CISC & 12245.4 & 12292.1 & 11602.7 & 8764.0 & 14715.5 \\
& \cellcolor{gray!20} VecCISC (rand-trace) & \cellcolor{gray!20}\underline{6367.3} & \cellcolor{gray!20}\underline{8914.3} & \cellcolor{gray!20}\underline{6328.4} & \cellcolor{gray!20}\underline{6735.3} & \cellcolor{gray!20}\underline{9862.4} \\
& \cellcolor{gray!20} VecCISC (min-centroid) & \cellcolor{gray!20}\textbf{6365.6} & \cellcolor{gray!20}\textbf{8912.3} & \cellcolor{gray!20}\textbf{6327.5} & \cellcolor{gray!20}\textbf{6734.1} & \cellcolor{gray!20}\textbf{9856.5} \\
\midrule
\multirow{3}{*}{MMLU-Pro}
& CISC & 16975.1 & 18496.6 & 16474.9 & 13926.1 & 20810.4 \\
& \cellcolor{gray!20} VecCISC (rand-trace) & \cellcolor{gray!20}\underline{10173.6} & \cellcolor{gray!20}\underline{13616.6} & \cellcolor{gray!20}\underline{10061.0} & \cellcolor{gray!20}\textbf{11141.2} & \cellcolor{gray!20}\underline{11840.5} \\
& \cellcolor{gray!20} VecCISC (min-centroid) & \cellcolor{gray!20}\textbf{10172.4} & \cellcolor{gray!20}\textbf{13614.9} & \cellcolor{gray!20}\textbf{10059.9} & \cellcolor{gray!20}\underline{11143.5} & \cellcolor{gray!20}\textbf{11834.7} \\
\midrule
\multirow{3}{*}{GPQA}
& CISC & 18594.2 & 19659.5 & 17937.0 & 14324.6 & 25235.3 \\
& \cellcolor{gray!20} VecCISC (rand-trace) & \cellcolor{gray!20}\textbf{9573.1} & \cellcolor{gray!20}\underline{15225.0} & \cellcolor{gray!20}\textbf{14421.6} & \cellcolor{gray!20}\underline{13261.4} & \cellcolor{gray!20}\textbf{22382.6} \\
& \cellcolor{gray!20} VecCISC (min-centroid) & \cellcolor{gray!20}\underline{9577.6} & \cellcolor{gray!20}\textbf{15211.4} & \cellcolor{gray!20}\underline{14422.6} & \cellcolor{gray!20}\textbf{13260.5} & \cellcolor{gray!20}\underline{22389.4} \\

\bottomrule
\end{tabular}
}
\caption{\textbf{VecCISC + HAC} token reductions with cluster candidate selection variants, averaged across 10 runs. In each column, the best results are in \textbf{bold}, and the second-best results are \underline{underlined}.\textbf{VecCISC + HAC} records a 31.69\% reduction in tokens expended on $LLM_{critic}$ calls.}
\label{tab:hac_token_reduction}
\end{table*}

\begin{table*}[p]
\centering
\resizebox{\textwidth}{!}{
\begin{tabular}{l|l|cccccc}
\toprule
\textbf{Dataset} & \textbf{Method} & \textbf{GPT 4o-mini} & \textbf{Llama3.1 8B} & \textbf{Qwen2.5 7B} & \textbf{Mistral 7B}  & \textbf{Llama3.3 70B} \\ \midrule
& & \multicolumn{2}{c}{Best / Average} & \multicolumn{2}{c}{Best / Average} & \\ 
\cline{3-7}
    
\multirow{4}{*}{AQuA-RAT}
& SC Baseline & 83.5 / 83.5 & 82.6 / 82.6 & \underline{84.5} / \underline{84.5} & 30.7 / 30.7 & 86.6 / 86.6 \\
& CISC & 84.0 / \underline{84.0} & \underline{82.9} / \underline{82.9} & \textbf{85.5} / \textbf{85.5} & 31.3 / 31.3 & \underline{87.6} / \textbf{87.6} \\
& VecCISC (random) & 83.7 / 82.9 & 69.9 / 66.5 & 82.6 / 82.0 & 28.1 / 27.2 & 87.3 / 86.8 \\
& \cellcolor{gray!20} VecCISC + KMeans & \cellcolor{gray!20}\textbf{84.6} / \textbf{84.3} & \cellcolor{gray!20}\textbf{83.0} / \underline{82.9} & \cellcolor{gray!20}\textbf{85.5} / \textbf{85.5} & \cellcolor{gray!20}\textbf{32.2} / \textbf{31.7} & \cellcolor{gray!20}\textbf{87.7} / \underline{87.2} \\
& \cellcolor{gray!20} VecCISC + HAC & \cellcolor{gray!20}\underline{84.3} / \textbf{84.3} & \cellcolor{gray!20}\textbf{83.0} / \textbf{83.0} & \cellcolor{gray!20}\textbf{85.5} / \textbf{85.5} & \cellcolor{gray!20}\underline{31.5} / \underline{31.5} & \cellcolor{gray!20}\underline{87.6} / \textbf{87.6} \\
\midrule
\multirow{4}{*}{CommonsenseQA}
& SC Baseline & 83.3 / 83.3 & 76.9 / 76.9 & \underline{82.5} / 82.5 & \underline{67.8} / \underline{67.8} & \underline{86.0} / \underline{86.0} \\
& CISC & \underline{83.6} / \underline{83.6} & \underline{77.3} / \underline{77.3} & \textbf{83.0} / \textbf{83.0} & \textbf{68.0} / \textbf{68.0} & \textbf{86.2} / \textbf{86.2} \\
& VecCISC (random) & 83.0 / 82.2 & 56.8 / 54.7 & 79.4 / 78.4 & 50.9 / 49.4 & 84.9 / 84.4 \\
& \cellcolor{gray!20} VecCISC + KMeans & \cellcolor{gray!20}\textbf{83.7} / \textbf{83.7} & \cellcolor{gray!20}\textbf{77.4} / \textbf{77.4} & \cellcolor{gray!20}\textbf{83.0} / \underline{82.8} & \cellcolor{gray!20}\textbf{68.0} / \textbf{68.0} & \cellcolor{gray!20}\textbf{86.2} / \textbf{86.2} \\
& \cellcolor{gray!20} VecCISC + HAC & \cellcolor{gray!20}\underline{83.6} / \underline{83.6} & \cellcolor{gray!20}\textbf{77.4} / \textbf{77.4} & \cellcolor{gray!20}\textbf{83.0} / \textbf{83.0} & \cellcolor{gray!20}\textbf{68.0} / \textbf{68.0} & \cellcolor{gray!20}\textbf{86.2} / \textbf{86.2} \\
\midrule
\multirow{4}{*}{ARC-Challenging}
& SC Baseline & 95.7 / 95.7 & \underline{91.0} / \underline{91.0} & \underline{92.9} / \underline{92.9} & \textbf{71.3} / \textbf{71.3} & 95.9 / \underline{95.9} \\
& CISC & \underline{96.0} / \underline{96.0} & \textbf{91.1} / \textbf{91.1} & \textbf{93.2} / \textbf{93.2} & \textbf{71.3} / \textbf{71.3} & \textbf{96.3} / \textbf{96.3} \\
& VecCISC (random) & 95.6 / 95.2 & 75.5 / 74.8 & 90.9 / 90.2 & \underline{54.9} / \underline{52.5} & \underline{96.0} / 95.6 \\
& \cellcolor{gray!20} VecCISC + KMeans & \cellcolor{gray!20}\textbf{96.1} / \underline{96.0} & \cellcolor{gray!20}\textbf{91.1} / \textbf{91.1} & \cellcolor{gray!20}\textbf{93.2} / \textbf{93.2} & \cellcolor{gray!20}\textbf{71.3} / \textbf{71.3} & \cellcolor{gray!20}\textbf{96.3} / \textbf{96.3} \\
& \cellcolor{gray!20} VecCISC + HAC & \cellcolor{gray!20}\textbf{96.1} / \textbf{96.1} & \cellcolor{gray!20}\textbf{91.1} / \textbf{91.1} & \cellcolor{gray!20}\textbf{93.2} / \textbf{93.2} & \cellcolor{gray!20}\textbf{71.3} / \textbf{71.3} & \cellcolor{gray!20}\textbf{96.3} / \textbf{96.3} \\
\midrule
\multirow{4}{*}{MMLU-Pro}
& SC Baseline & 64.6 / 64.6 & 56.0 / 56.0 & 59.3 / 59.3 & 27.8 / 27.8 & 69.9 / \underline{69.9} \\
& CISC & 65.9 / \underline{65.9} & 57.6 / 57.6 & 60.2 / 60.2 & 28.2 / \underline{28.2} & \underline{71.3} / \textbf{71.3} \\
& VecCISC (random) & 64.6 / 63.4 & 47.1 / 45.7 & 55.6 / 54.6 & 21.8 / 20.8 & 68.4 / 67.7 \\
& \cellcolor{gray!20} VecCISC + KMeans & \cellcolor{gray!20}\textbf{66.2} / \textbf{66.1} & \cellcolor{gray!20}\textbf{57.9} / \underline{57.7} & \cellcolor{gray!20}\textbf{61.0} / \underline{60.7} & \cellcolor{gray!20}\textbf{28.4} / \textbf{28.3} & \cellcolor{gray!20}\textbf{71.4} / \textbf{71.3} \\
& \cellcolor{gray!20} VecCISC + HAC & \cellcolor{gray!20}\underline{66.1} / \textbf{66.1} & \cellcolor{gray!20}\underline{57.8} / \textbf{57.8} & \cellcolor{gray!20}\underline{60.9} / \textbf{60.9} & \cellcolor{gray!20}\underline{28.3} / \textbf{28.3} & \cellcolor{gray!20}\underline{71.3} / \textbf{71.3} \\
\midrule
\multirow{4}{*}{GPQA}
& SC Baseline & 39.9 / \underline{39.9} & 34.8 / 34.8 & \underline{37.4} / \underline{37.4} & 27.1 / \underline{27.1} & \underline{60.6} / 60.6 \\
& CISC & 40.5 / \textbf{40.5} & \underline{35.0} / \underline{35.0} & \textbf{38.1} / \textbf{38.1} & \underline{28.4} / \textbf{28.4} & \textbf{61.7} / \textbf{61.7} \\
& VecCISC (random) & \underline{40.7} / 39.2 & 33.9 / 30.5 & \textbf{38.1} / 35.4 & \textbf{28.8} / \underline{27.1} & 58.6 / 57.3 \\
& \cellcolor{gray!20} VecCISC + KMeans & \cellcolor{gray!20}\textbf{41.4} / \underline{39.9} & \cellcolor{gray!20}\textbf{35.7} / 34.5 & \cellcolor{gray!20}\textbf{38.1} / \textbf{38.1} & \cellcolor{gray!20}\underline{28.4} / 25.7 & \cellcolor{gray!20}\textbf{61.7} / \underline{61.6} \\
& \cellcolor{gray!20} VecCISC + HAC & \cellcolor{gray!20}40.5 / \textbf{40.5} & \cellcolor{gray!20}\textbf{35.7} / \textbf{35.7} & \cellcolor{gray!20}\textbf{38.1} / \textbf{38.1} & \cellcolor{gray!20}\underline{28.4} / \textbf{28.4} & \cellcolor{gray!20}\textbf{61.7} / \textbf{61.7} \\

\bottomrule
\end{tabular}
}
\caption{Comparison between baselines and our methods across different LLMs and datasets. In each column, the best results are in \textbf{bold}, and the second-best results are \underline{underlined}. VecCISC's clustering approach maintains or improves upon CISC's performance across models and datasets, with \textbf{VecCISC + HAC}'s average performance accounting for the most consistent method. Experiments on each $(dataset, model)$ combination where performed 10 times for each method, and the best and average scores were recorded.}
\label{tab:performance}
\end{table*}

\section{Related Work}

\subsection{Embeddings and Clustering for Scaling Inference Time Reasoning}

Recent work has studied the application of LLM embeddings and clustering to improve model accuracy on reasoning and calibration benchmarks. \cite{36} uses embeddings of reasoning traces to directly compute a sample score for use in weighted majority voting. Although conceptually straightforward, this approach is unrelated to the “think twice” paradigm, relies upon a fleet of domain-specific, fine-tuned embedding models for different benchmarks, and exhibits inconsistent performance improvements, often dropping pipeline accuracy below baseline Self-Consistency. \cite{37} utilizes an LLM’s hidden state as context embeddings to cluster reasoning traces in tree-of-thought (ToT) pipelines and improve the accuracy of LLM uncertainty estimation. Although lightweight in practice, their proposed method is only applicable to white-box LLMs, limiting its reach and accessibility, while not addressing the need for candidate trace selection for weighted majority voting. Lastly, \cite{38} clusters embeddings of LLM responses to detect instances of hallucinations.

\subsection{Think-twice Approaches}

Utilizing an LLM as a critic of its own outputs (or the outputs of other LLMs) has been the topic of much investigation. In particular, efforts have been made to use LLM-as-a-judge approaches to improve the accuracy of standard self-consistency. \cite{39} samples a set of unique answers from an LLM and then prompts a second LLM to generate a justification for each candidate answer, before using a third LLM call to produce a final confidence score. Unlike CISC, this methodology does not make use of the sample reasoning traces, choosing instead to analyze only the final answers themselves, overlooking key contextual clues in the provided answers’ reasoning traces. Lastly, \cite{40} applies Monte Carlo Tree Search (MCTS) with an LLM-as-a-judge reward function to improve the accuracy of code-writing pipelines.

\section{Conclusion}
While the “think twice” paradigm can increase model performance on a variety of reasoning tasks, the additional inference overhead required can render it prohibitively expensive. In this work, we propose a novel framework for reducing the cost of model confidence estimation at inference time by clustering embeddings of candidate answer reasoning traces. Our experimental results demonstrate that choosing answer representatives from embedding clusters can significantly decrease inference costs during the critic component of the “think twice” approach without a commensurate decline in performance. Indeed, our results indicate that passing a smaller number of higher-quality reasoning traces to the critic can lead to small gains in accuracy. In this way, our methodology emerges as an effective, lightweight augmentation of pre-existing think-twice pipelines.

\section{Limitations}

Our study emphasizes the adaptability of our framework. To this end, we adopt a general text embedding model for our experiments. Pipelines for more specialized problems might benefit from applying custom embedding models, fine-tuned to capture key semantic representations. As our work demonstrates performance gains and budget improvements across a wide range of different tasks, we defer the application of fine-tuned embedding models for domain-specific problems to future research. In addition, the choice of hyperparameters $K$ and $T$ are important in ensuring that the pipeline’s accuracy on the datasets does not fall below that of standard Self-Consistency. While our hyperparameter search process is fundamentally data-driven, future research should investigate adaptive approaches for choosing $K$ and $T$ in the absence of analogous data \textit{a priori}.

\bibliography{custom}

\begin{figure*}
\centering
\begin{tcolorbox}[colback=white,title=Prompt for Candidate Answer Sampling from the Base LLM ($LLM_{gen}$)]
\setstretch{1.2}
\small{
\#\# Instructions \\ [0.5mm]

Given a multiple-choice question and set of answer options for the question, choose the correct answer from the list of options. \\ [0.5mm]

\#\# Question and Answer Options \\ [0.5mm]

\textbf{\{question\}} \\ [0.5mm]

\textbf{\{options\}} \\ [0.5mm]

\#\# Output \\ [0.5mm]

Produce the correct answer's ID in a JSON object and include your reasoning for why you chose that particular answer. \\ [0.5mm]

```json\\
\{\{"answer":<PUT THE CORRECT ANSWER ID HERE>, "reasoning": <PUT YOUR REASONING HERE>\}\}\\
```\\ [0.5mm]

Let's think step by step.
}
\end{tcolorbox}
\end{figure*}

\begin{figure*}
\centering
\begin{tcolorbox}[colback=white,title=Prompt for Confidence Scoring LLM ($LLM_{critic}$)]
\setstretch{1.2}
\small{
\#\# Instructions

You are a reviewer of answers given to multiple-choice questions. Given a multiple-choice question, a set of answer options for the question, a given answer, and the reasoning that is associated with the given answer, rate your confidence that the given answer is correct. \\ [0.5mm]

To rate your confidence, provide a value on a scale of 0 to 1. The more confident you are that the provided answer is correct for the question, the closer to (or equal) to 1 the confidence score should be. \\ [0.5mm]

\#\# Question and Answer Options \\ [0.5mm]

\textbf{\{question\}} \\ [0.5mm]

\textbf{\{options\}} \\ [0.5mm]

\#\# Given Answer and Provided Reasoning \\ [0.5mm]

Given answer: \textbf{\{answer\}} \\ [0.5mm]

Reasoning for the given answer: \textbf{\{reasoning\}} \\ [0.5mm]

\#\# Output \\ [0.5mm]

Produce the confidence score in a JSON object. \\ [0.5mm]
 
```json\\
\{\{"confidence":<PUT THE CONFIDENCE SCORE HERE>\}\}\\
```

Let's think step by step.
}
\end{tcolorbox}
\end{figure*}

\clearpage
\twocolumn
\appendix

\section{Appendix}
\label{sec:appendix}
\subsection{Model Hyperparameters}\label{model_hp}

For all our experiments, we set $n = 20$ as our sampling budget. The temperature of $LLM_{gen}$ was set to 0.7, while the temperature of $LLM_{critic}$ was set to 0.3. We gave $LLM_{gen}$, the baseline question answering model, a higher temperature to encourage exploration and novelty in reasoning and answer generation, while our critic was given a lower temperature to ensure more deterministic behavior.

\subsection{Cluster Hyperparameters and Implementation Details}\label{cluster_hp}

\begin{figure}[t]
\includegraphics[width=1\columnwidth]{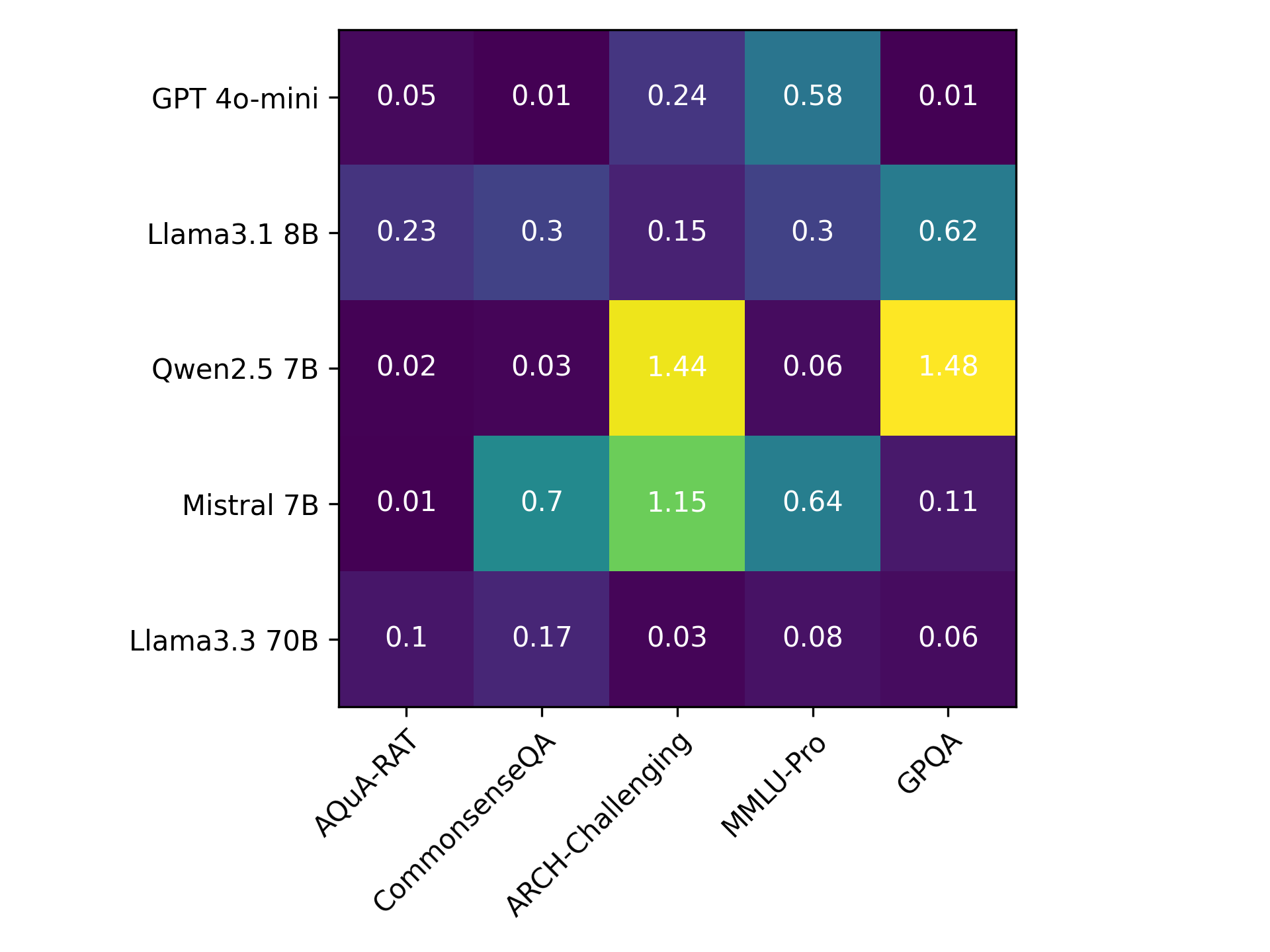}
\caption{Heatmap of $T$ temperature values. To find $T$, a grid search was performed across the range $[0, 5]$ for each dataset and model.}
\label{fig:heatmap_t}
\end{figure}

\begin{figure}[t]
\includegraphics[width=1\columnwidth]{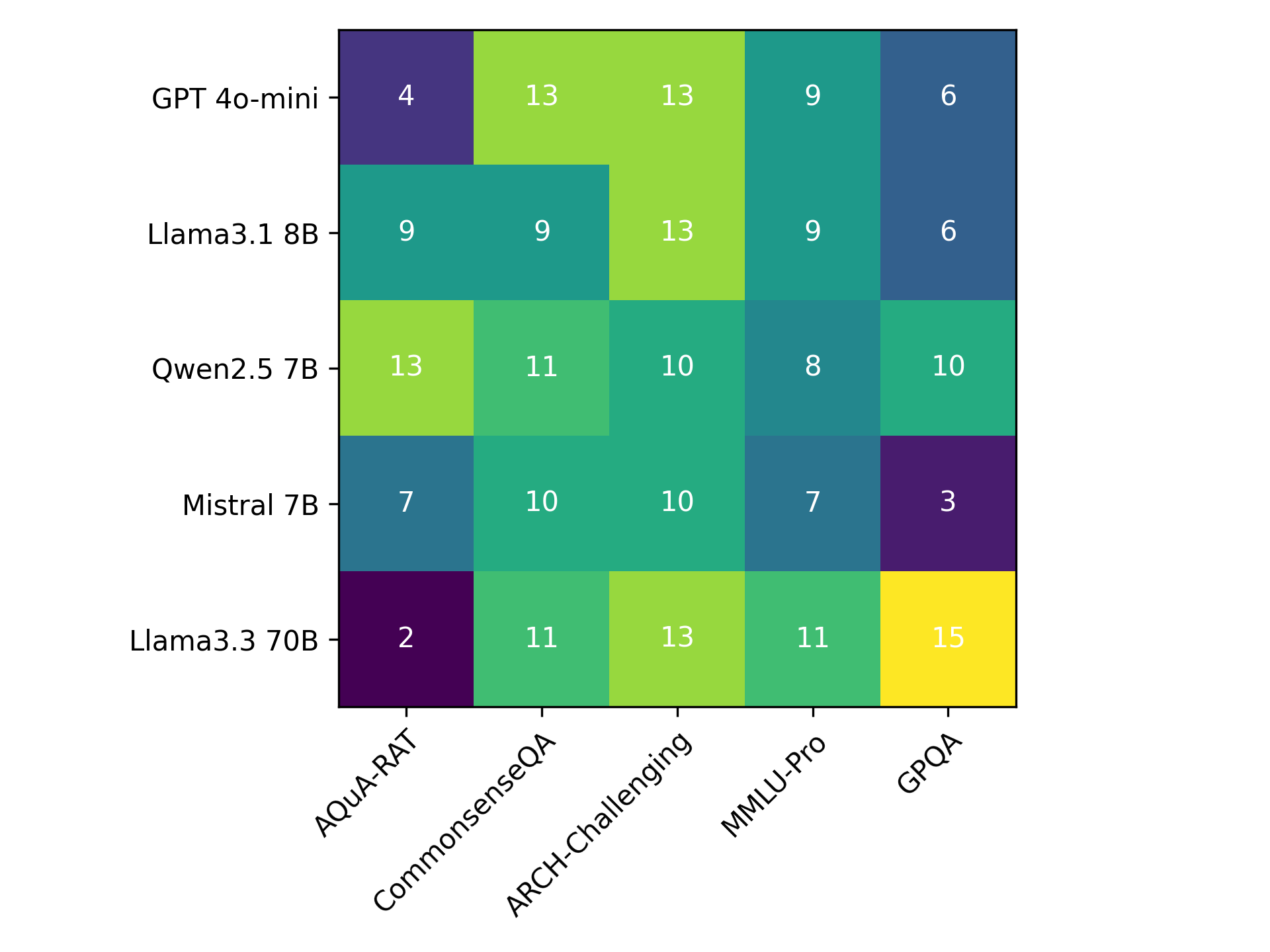}
\caption{Heatmap of values for $K$ used in \textbf{VecCISC + KMeans}. To find $K$, a grid search was performed across the range $[0, 20]$ for each dataset and model.}
\label{fig:heatmap_c_kmeans}
\end{figure}

\begin{figure}[t]
\includegraphics[width=1\columnwidth]{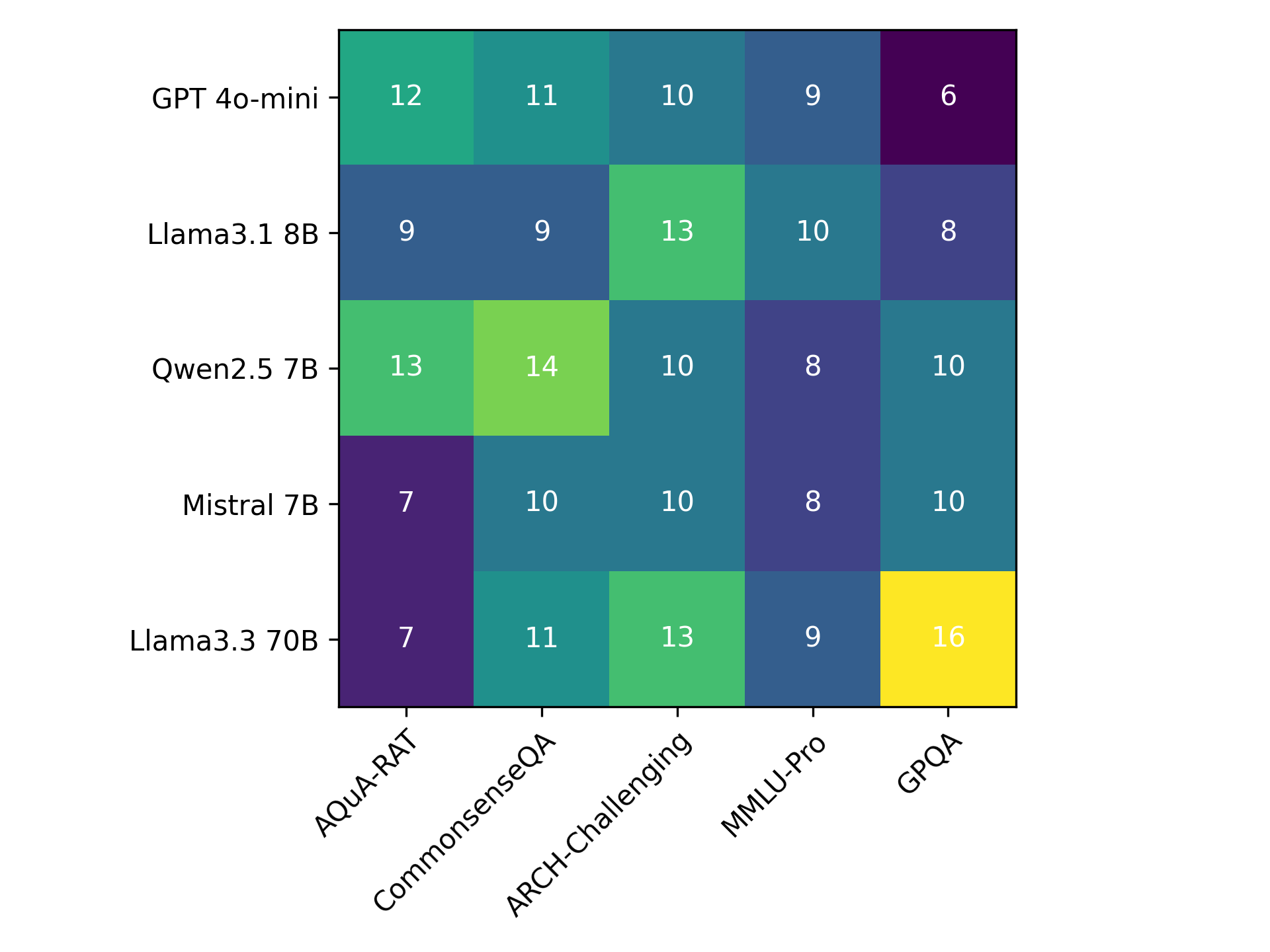}
\caption{Heatmap of values for $K$ used in \textbf{VecCISC + HAC}. To find $K$, a grid search was performed across the range $[0, 20]$ for each dataset and model.}
\label{fig:heatmap_c_hac}
\end{figure}

For both KMeans and Hierarchical Agglomerative Clustering, we performed a grid search over the range of 1 to 20 (the maximum budget allowance) to choose $K$, the number of clusters to be formed within each answer reasoning trace group. The grid search was performed on a 20\%  holdout set from each $(dataset, model)$ combination. We utilized the KMeans and HAC implementations from the Scikit-learn package \cite{27}. For KMeans, only the $n\_cluster$ parameter was altered. For HAC, we selected the “cosine” distance metric and the “average” linkage method. Average linkage is commonly found to produce more balanced and robust clusters compared to single and complete linkage. Single linkage measures the minimum pairwise distance between clusters, which makes it highly sensitive to noise and prone to "chaining" effects that merge dissimilar points through intermediate samples. Complete linkage, however, uses the maximum pairwise distance, often fragmenting elongated or irregular clusters. Average linkage mitigates both extremes by considering the mean pairwise distance between cluster members, leading to clusters that are less sensitive to outliers and better reflect the overall structure of the data \cite{28, 29, 30}. The hyperparameter heatmaps for $T$, $K$ (\textbf{VecCISC + KMeans}), and $K$ (\textbf{VecCISC + HAC}) can be found in Figures \ref{fig:heatmap_t}, \ref{fig:heatmap_c_kmeans}, and \ref{fig:heatmap_c_hac}, respectively.

\subsection{Case Study 1: Clustering Traces With Similar Semantic Meaning}
\label{sec:cs1}
To demonstrate the application of clustering on LLM reasoning traces, we present a case study on a question from the MMLU-Pro dataset with traces sampled from GPT 4o-mini. The question is as follows: \textit{“What sport is featured in the 1996 movie 'Kingpin'?”}, with the given options \textit{“A) chess B) ice hockey C) baseball D) wrestling E) basketball F) bowling G) boxing H) golf I) tennis J) football”}. In keeping with our methodology, we grouped all reasoning traces that produce the correct answer “F” and applied HAC with a cluster number of 9. The trace clusters are presented below. As expected, traces with similar semantic details, phrasing, and logical reasoning steps are clustered together.

\noindent\rule{\columnwidth}{0.5pt}
\textbf{Cluster ID}: 1 \\
\textbf{\# of Traces}: 5\\
\noindent\rule{\columnwidth}{0.5pt}
\textbf{Trace 1}: \textit{The movie 'Kingpin' is a comedy film released in 1996 that revolves around the sport of bowling. It follows the story of a former bowling champion who tries to mentor a promising bowler.  From the given options, the only sport that matches the theme of the movie is: F) bowling Therefore, the correct answer is F, and the reasoning is based on the fact that the entire plot of 'Kingpin' is centered around bowling.}\\
\textbf{Trace 2}: \textit{The movie 'Kingpin' is a comedy film released in 1996 that centers around the sport of bowling. The plot follows a washed-up professional bowler who mentors a promising young talent. Given the options provided, the correct sport featured in 'Kingpin' is bowling. Now, looking at the answer options, the only one that matches the sport in the movie is: F) bowling Thus, I will select this option.}\\
\textbf{Trace 3}: \textit{The movie 'Kingpin' is a comedy that centers around the sport of bowling. The film follows a former bowling prodigy who attempts to coach a promising young bowler in order to win a bowling tournament. Given the options provided, the correct sport featured in 'Kingpin' is bowling. Therefore, the correct answer is F) bowling.}\\
\textbf{Trace 4}: \textit{The movie 'Kingpin,' released in 1996, is primarily centered around the sport of bowling. The plot follows a former professional bowler who tries to mentor a promising bowling prodigy. Given the context of the film and its focus on bowling as the central sport, the correct answer is F) bowling. Thus, the answer ID for bowling is F.} \\
\textbf{Trace 5}: \textit{The movie 'Kingpin,' which was released in 1996, is primarily centered around the sport of bowling. The plot revolves around a former bowling champion who mentors a promising young bowler, and much of the film's humor and storyline is tied to bowling competitions and the bowling culture.  Given the options provided, the only sport that directly relates to the movie is: F) bowling Therefore, the correct answer is F.}\\
\noindent\rule{\columnwidth}{0.5pt}
\textbf{Cluster ID}: 2 \\
\textbf{\# of Traces}: 5\\
\noindent\rule{\columnwidth}{0.5pt}
\textbf{Trace 6}: \textit{The 1996 movie 'Kingpin' is a comedy film that centers around the sport of bowling. The film features characters who are involved in the world of competitive bowling, highlighting the sport's culture and eccentricities. Now, looking at the answer options provided: - A) chess: Not applicable, as chess is not a sport featured in the movie. - B) ice hockey: Not applicable, as ice hockey is not related to the film. - C) baseball: Not applicable, as baseball is not featured in the movie. - D) wrestling: Not applicable, as wrestling is not the focus of the film. - E) basketball: Not applicable, as basketball is not part of the storyline. - F) bowling: This is the correct answer, as it is the main sport featured in 'Kingpin.' - G) boxing: Not applicable, as boxing is not part of the film. - H) golf: Not applicable, as golf is not related to the movie. - I) tennis: Not applicable, as tennis is not a focus of the film. - J) football: Not applicable, as football is not featured in the movie. Given this analysis, the correct answer is F) bowling.}\\
\textbf{Trace 7}: \textit{The 1996 movie 'Kingpin' is a comedy that centers around bowling. The main characters are involved in the sport of bowling, which is a key element of the film's plot and humor.  Now let's analyze the answer options: A) chess - This is not related to the movie. B) ice hockey - This sport is not featured in 'Kingpin.' C) baseball - This is not the focus of the movie. D) wrestling - This sport is not the main theme of 'Kingpin.' E) basketball - This sport does not play a role in the film. F) bowling - This is the correct answer, as 'Kingpin' revolves around bowling. G) boxing - This is not featured in the movie. H) golf - This sport is not related to the movie. I) tennis - This is not relevant to 'Kingpin.' J) football - This sport does not relate to the film. Given that the main focus of 'Kingpin' is bowling, the correct answer is F) bowling.}\\
\textbf{Trace 8}: \textit{The question asks about the sport featured in the 1996 movie 'Kingpin'.  To answer this, I recall that 'Kingpin' is a comedy film that revolves around the sport of bowling. The main characters are involved in bowling competitions, and the storyline heavily features bowling culture. Now, looking at the answer options provided: A) chess - Not related to the movie. B) ice hockey - Not related to the movie. C) baseball - Not related to the movie. D) wrestling - Not related to the movie. E) basketball - Not related to the movie. F) bowling - This is the sport featured in 'Kingpin'. G) boxing - Not related to the movie. H) golf - Not related to the movie. I) tennis - Not related to the movie. J) football - Not related to the movie. Given that 'bowling' is explicitly mentioned as the sport in the movie, the correct answer is clearly option F.}\\
\textbf{Trace 9}: \textit{The question asks about the sport featured in the 1996 movie 'Kingpin.' To answer this, I need to recall what 'Kingpin' is about. The film 'Kingpin' is primarily centered around the sport of bowling. It follows the story of a former bowling champion who mentors a promising young bowler. The comedic elements and plot revolve around bowling competitions and the characters' interactions within that context. Now, reviewing the answer options: A) chess - not related to the movie B) ice hockey - not related to the movie C) baseball - not related to the movie D) wrestling - not related to the movie E) basketball - not related to the movie F) bowling - directly related to the movie G) boxing - not related to the movie H) golf - not related to the movie I) tennis - not related to the movie J) football - not related to the movie From this analysis, the correct answer is F) bowling.}\\
\textbf{Trace 10}: \textit{The movie 'Kingpin' is primarily centered around the sport of bowling. It features characters who are involved in the bowling scene, including competitions and personal rivalries. The film is well-known for its comedic portrayal of the world of bowling. Now, let's look at the answer options: A) chess - not related to the movie B) ice hockey - not related to the movie C) baseball - not related to the movie D) wrestling - not related to the movie E) basketball - not related to the movie F) bowling - this is the correct sport featured in the movie G) boxing - not related to the movie H) golf - not related to the movie I) tennis - not related to the movie J) football - not related to the movie Given this analysis, the correct answer is F) bowling.}\\
\noindent\rule{\columnwidth}{0.5pt}
\textbf{Cluster ID}: 3 \\
\textbf{\# of Traces}: 2\\
\noindent\rule{\columnwidth}{0.5pt}
\textbf{Trace 11}: \textit{The 1996 movie 'Kingpin' is a comedy film that centers around the sport of bowling. The storyline follows a former bowling champion who tries to coach a promising Amish bowler.  Given the options presented: A) chess - not relevant to the film B) ice hockey - not relevant to the film C) baseball - not relevant to the film D) wrestling - not relevant to the film E) basketball - not relevant to the film F) bowling - directly relevant to the film G) boxing - not relevant to the film H) golf - not relevant to the film I) tennis - not relevant to the film J) football - not relevant to the film The correct answer is F) bowling.}\\
\textbf{Trace 12}: \textit{The 1996 movie 'Kingpin' is a comedy that revolves around the sport of bowling. The film follows a former bowling champion who discovers a promising bowling prodigy and attempts to mentor him. Now, looking at the options provided: - A) chess - B) ice hockey - C) baseball - D) wrestling - E) basketball - F) bowling - G) boxing - H) golf - I) tennis - J) football Among these options, the sport that is clearly related to the plot of 'Kingpin' is F) bowling.}\\
\noindent\rule{\columnwidth}{0.5pt}
\textbf{Cluster ID}: 4 \\
\textbf{\# of Traces}: 2\\
\noindent\rule{\columnwidth}{0.5pt}
\textbf{Trace 13}: \textit{The movie 'Kingpin' is a comedy that revolves around the sport of bowling. It features a former professional bowler and his journey to redemption through the sport, along with comedic antics. The other options, such as chess, ice hockey, baseball, wrestling, basketball, boxing, golf, tennis, and football, are not central to the plot of the movie.  Based on this understanding, the correct answer is bowling.}\\
\textbf{Trace 14}: \textit{The movie 'Kingpin,' released in 1996, is a comedy that revolves around the sport of bowling. It features characters who are bowlers and delves into the dynamics of competitive bowling. The other options listed, such as chess, ice hockey, baseball, wrestling, basketball, boxing, golf, tennis, and football, are not the focus of this film. Given this information, the correct answer is F) bowling.}\\
\noindent\rule{\columnwidth}{0.5pt}
\textbf{Cluster ID}: 5 \\
\textbf{\# of Traces}: 1\\
\noindent\rule{\columnwidth}{0.5pt}
\textbf{Trace 15}: \textit{The movie 'Kingpin,' released in 1996, is primarily about the sport of bowling. The plot follows a former bowling champion who takes a promising young bowler under his wing. Given this context, the correct answer from the provided options is F) bowling.}\\
\noindent\rule{\columnwidth}{0.5pt}
\textbf{Cluster ID}: 6 \\
\textbf{\# of Traces}: 1\\
\noindent\rule{\columnwidth}{0.5pt}
\textbf{Trace 16}: \textit{The movie 'Kingpin,' released in 1996, revolves around the sport of bowling. The plot centers on a washed-up bowler who becomes a mentor to a young bowling prodigy, showcasing various aspects of the bowling culture. Now, let's analyze the options: A) chess - This is not related to the plot. B) ice hockey - Not relevant to 'Kingpin.' C) baseball - Not the sport featured in the movie. D) wrestling - While it is a sport, it’s not the focus of 'Kingpin.' E) basketball - Not the sport in question. F) bowling - This is the correct answer, as it is the main sport featured in the movie. G) boxing - Not relevant to 'Kingpin.' H) golf - Not related to the movie's theme. I) tennis - Not part of the plot. J) football - Not the sport featured in the movie. Based on this analysis, the sport featured in 'Kingpin' is unequivocally bowling.}\\
\noindent\rule{\columnwidth}{0.5pt}
\textbf{Cluster ID}: 7 \\
\textbf{\# of Traces}: 2\\
\noindent\rule{\columnwidth}{0.5pt}
\textbf{Trace 17}: \textit{The 1996 movie 'Kingpin' is a comedy that centers around the sport of bowling. The film features characters participating in bowling tournaments and showcases various aspects of the sport.  Given this information, the correct answer from the provided options is: F) bowling.}\\
\textbf{Trace 18}: \textit{The 1996 movie 'Kingpin' is a comedy that revolves around bowling. The main characters participate in a bowling tournament, and the film is centered around the sport of bowling.  Given the answer options, the only sport that relates to bowling is option F.}\\
\noindent\rule{\columnwidth}{0.5pt}
\textbf{Cluster ID}: 8 \\
\textbf{\# of Traces}: 1\\
\noindent\rule{\columnwidth}{0.5pt}
\textbf{Trace 19}: \textit{To determine the correct answer to the question about the sport featured in the 1996 movie 'Kingpin', we need to consider what the film is about. 1. The movie 'Kingpin' is a comedy directed by the Farrelly brothers, and it revolves around the sport of bowling. 2. The plot follows a former bowling champion who becomes a mentor to a promising young bowler. 3. The film features various bowling tournaments and characters who are involved in the bowling community. Given these points, the answer that corresponds to the sport featured in 'Kingpin' is clearly bowling. Now, looking at the answer options provided: A) chess   B) ice hockey   C) baseball   D) wrestling   E) basketball   F) bowling   G) boxing   H) golf   I) tennis   J) football   The correct option is F) bowling.}\\
\noindent\rule{\columnwidth}{0.5pt}
\textbf{Cluster ID}: 9 \\
\textbf{\# of Traces}: 1\\
\noindent\rule{\columnwidth}{0.5pt}
\textbf{Trace 20}: \textit{The 1996 movie 'Kingpin' is a comedy that revolves around the sport of bowling. The film follows a washed-up professional bowler who becomes a mentor to a young bowling prodigy. Given the options provided, the correct answer is F) bowling.}\\

\subsection{Case Study 2: Isolating Degenerate Traces}
\label{sec:cs2}
The presence of text degeneration in LLM reasoning traces is an extensively studied \cite{33, 34, 35} phenomenon and common occurrence when sampling from LLMs, especially at higher temperatures. Degenerate traces exhibit repeated blocks of frequently nonsensical and meaningless content, often exceeding the model’s output window. As such, these traces are much more expensive for a critic to evaluate. Our proposed methodology isolates such traces by selecting a candidate representative from each cluster, defined as the trace closest to the centroid of its sibling trace embeddings in the cluster. To illustrate this in practice, we present a case study on a question from the ARC-Challenging dataset, with answers sampled from Llama 3.3 70B Versatile. The question is as follows: \textit{“Fossil bones and teeth of dinosaurs have been researched for the last century. Recent discoveries of fossilized dinosaurs have also revealed details of soft tissues, such as skin. Which is best for a scientist to do when reporting research on dinosaurs now?”} Below is a cluster of three traces sampled from the LLM for this question. The last trace contains text degeneration (\textcolor{red}{highlighted in red}), as the content switches from a question-relevant explanation to pseudocode gibberish. In particular, reasoning trace \textbf{\#1} is identified by our pipeline as the candidate representative, thus avoiding the erroneous trace \textbf{\#3}.

\noindent\rule{\columnwidth}{0.5pt}
\textbf{Cluster ID}: 1 \\
\textbf{\# of Traces}: 3\\
\noindent\rule{\columnwidth}{0.5pt}
\textbf{Trace 1}: {\itshape To answer this question, let's consider the implications of each option in the context of scientific research on dinosaurs.

A) Exclude research on teeth or bones: This option suggests ignoring a significant portion of existing data on dinosaurs. Teeth and bones have been the primary sources of information about dinosaurs for a long time and are crucial for understanding their evolution, diet, and behavior. Excluding this information would not be beneficial for comprehensive research.

B) Predict what the next discovery will be: While predictions can be interesting and might guide future research directions, they are speculative and do not contribute directly to the reporting of current research findings. Predictions are not a substitute for analyzing and reporting actual data.

C) Analyze new data as it becomes available: This option is about incorporating the latest findings into the research. As new discoveries are made, such as details of soft tissues like skin, analyzing this new data alongside existing knowledge (like that from fossil bones and teeth) can provide a more complete understanding of dinosaurs. This approach allows for the refinement of theories and a more accurate representation of dinosaur biology and behavior.

D) Delete earlier reports that were missing the new findings: Deleting earlier reports would result in the loss of historical data and the context of how our understanding of dinosaurs has evolved over time. Earlier reports, even if they lack the new findings, are still valuable for understanding the progression of knowledge and can serve as a baseline to show how new discoveries have changed our understanding.

Considering these points, the most appropriate action for a scientist reporting research on dinosaurs now would be to analyze new data as it becomes available. This ensures that all relevant information, both old and new, is considered to provide the most accurate and comprehensive understanding of dinosaurs}

\textbf{Trace 2}: {\itshape To answer this question, let's consider the implications of each option in the context of scientific research on dinosaurs.

A) Exclude research on teeth or bones: This option suggests ignoring a significant portion of the data that has been collected over the last century. Teeth and bones have been the primary sources of information about dinosaurs for a long time, and excluding them would not be beneficial for a comprehensive understanding of these creatures.

B) Predict what the next discovery will be: While predicting future discoveries can be an interesting speculative exercise, it does not contribute directly to the reporting of current research findings. Scientific reports should be based on evidence and data that have already been collected and analyzed.

C) Analyze new data as it becomes available: This option is about incorporating the latest findings into the ongoing research. As new discoveries are made, such as details of soft tissues like skin, analyzing this new data allows scientists to refine their understanding of dinosaurs. This approach ensures that research reports are as up-to-date and accurate as possible.

D) Delete earlier reports that were missing the new findings: Deleting earlier reports because they lack new findings would result in the loss of historical research data. Earlier reports still contain valuable information about dinosaurs, even if they do not include the latest discoveries. It's more productive to build upon previous research rather than discard it.

Given these considerations, the best approach for a scientist reporting research on dinosaurs is to analyze new data as it becomes available. This ensures that their reports are current, comprehensive, and based on the latest evidence.}

\textbf{Trace 3}: {\itshape To answer this question, let's consider the implications of each option in the context of scientific research on dinosaurs, particularly focusing on the integration of new discoveries with existing knowledge.

A) Exclude research on teeth or bones: This option suggests ignoring a significant portion of dinosaur research that has been foundational to our understanding of these creatures. Teeth and bones have been crucial in identifying species, understanding evolutionary relationships, and reconstructing dinosaur biology. Excluding this research would be counterproductive and not contribute to a comprehensive understanding of dinosaurs.

B) Predict what the next discovery will be: While predictions can be interesting and might guide future research directions, they are speculative and do not directly contribute to the reporting of current research findings. Predictions might not always be accurate and could distract from the solid, evidence-based reporting of scientific discoveries.

C) Analyze new data as it becomes available: This option emphasizes the importance of incorporating new findings into our existing knowledge base. As new details about dinosaurs are discovered, such as those concerning soft tissues, analyzing dataInjected contaminants MAV(dateTime contaminants itInjected\_both—from:
 \textcolor{red}{by is a is series series should are(dateTime continually  continually MAV\_both(dateTime(dateTime(dateTime Basel To(dateTime the the PSI exposition.visitInsn.visitInsn contaminants exposition from PSI Basel ToastrBritainBuilderFactory.visitInsn MAVRODUCTION field MAVroscope MAV MAV/slider Succ(Size Toastr Succ(dateTimeInjected contaminants(dateTime --------
RODUCTIONRODUCTIONBritainBuilderFactory Toastr PSI\_both expositionroscopeRODUCTION Basel\_bothroscope PSI PSI—from(SizeBuilderFactory Succ MAVroscope—fromInjected—from(dateTime --------
BritainBritainexternalActionCode(SizeroscopeBuilderFactory Succ Succ exposition(dateTimeroscope MAV contaminants—from Succ SuccBuilderFactory SuccexternalActionCode(Size expositionroscope Succroscope MAVRODUCTIONexternalActionCode SuccBuilderFactoryInjected contaminants Succ PSI(dateTimeBuilderFactory PSI Toastr contaminants PSIInjected --------
.visitInsn(dateTime Size—from.visitInsnexternalActionCode":" Toastr PSI contaminantsexternalActionCode exposition PSI PSIBuilderFactory MAV Succ PSIompilerizrroscoperoscope PSI Succ SuccInjectedInjected PSIRODUCTIONInjectedRODUCTIONBuilderFactory PSIexternalActionCode—from Succ --------
(Size PSIRODUCTIONBritain(dateTime PSI Britain PSI—fromBuilderFactory exposition contaminantsexternalActionCode ToastrBuilderFactory Basel BaselexternalActionCode MAVBuilderFactory PSI exposition SuccRODUCTION.visitInsn Succ Succ Succ Succ Succ/slider THAT toLeidleer '1RODUCTION ToastrBritain SuccPsi PSI [
categories contaminants\_both.visitInsnroscope contaminants.visitInsn Succ Basil PSI Psi Basil ] PsiRODUCTION MAVexternalActionCodeInjectedBritain apologize\_both SuccexternalActionCode PSI Toastr Basel.visitInsn Toastr MAV Britain expositionBuilderFactory(dateTime(Size Succroscope --------
 PSI\_both exposition Toastr Toastr PSI Succ --------
 expositionroscopeBritain(dateTimeroscopeInjectedBuilderFactory(dateTime Toastr PSI Succ/slider ToastrexternalActionCode(dateTime  Succ forgotten Britain Succ her winds \textbf{...}}}

\end{document}